\documentclass[sigconf]{acmart}
\renewcommand\footnotetextcopyrightpermission[1]{} 
\usepackage{booktabs} 
\usepackage{hyperref}
\usepackage{subfigure}
\usepackage{amsmath}
\usepackage{url}
\usepackage{bm}
\usepackage{algorithm}
\usepackage{algorithmic}
\usepackage[export]{adjustbox}

\setcopyright{rightsretained}

\begin{document}
\title{Asynchronous Stochastic Variational Inference}

\author{Saad~Mohamad}
\affiliation{%
  \institution{Department of Computing, Bournemouth University}
  \city{Poole}
  \country{UK}
}
\email{smohamad@bournemouth.ac.uk}

\author{Abdelhamid~Bouchachia}
\affiliation{%
    \institution{Department of Computing, Bournemouth University}
  \city{Poole}
  \country{UK}
}
\email{abouchachia@bournemouth.ac.uk}

\author{Moamar~Sayed-Mouchaweh}
\affiliation{%
  \institution{Department of Informatics and Automatics, IMT Lille Douai}
  \city{Douai}
  \country{France}}
\email{moamar.sayed-mouchaweh@imt-lille-douai.fr}

\renewcommand{\shortauthors}{S. Mohamad et al.}

\begin{abstract}
Stochastic variational inference (SVI) employs stochastic optimization to scale up Bayesian computation to massive data. Since SVI is at its core a stochastic gradient-based algorithm, horizontal parallelism can be harnessed to allow larger scale inference. We propose a lock-free parallel implementation for SVI which allows distributed computations over multiple slaves in an asynchronous style. We show that our implementation  leads to linear speed-up while guaranteeing an asymptotic ergodic convergence rate $O(1/\sqrt{T})$ given that the number of slaves is bounded by $\sqrt{T}$ ($T$ is the total number of iterations). The implementation is done in a high-performance computing (HPC) environment using message passing interface (MPI) for python (MPI4py). The extensive empirical evaluation shows that our parallel SVI is lossless, performing comparably well to its counterpart serial SVI with linear speed-up.\vspace{-0.2cm}
\end{abstract}


\maketitle
\section{Introduction}\label{sec:introduction}
Probabilistic models with latent variables have grown into a backbone in many modern machine learning applications such as text analysis, computer vision, time series analysis, network modelling, and others. The main challenge in such models is to compute the posterior distribution over some hidden variables encoding hidden structure in the observed data. Generally, computing the posterior is intractable and approximation is required. Markov chain Monte Carlo (MCMC) sampling has been the dominant paradigm for posterior computation. It constructs a Markov chain on the hidden variables whose stationary distribution is the desired posterior. Hence, the approximation is based on sampling for a long time to (hopefully) collect samples from the posterior~\cite{andrieu2003introduction}.

 Recently, variational inference (VI) has  become  widely used as a deterministic alternative approach to MCMC sampling. In general, VI tends to be faster than MCMC which makes it more suitable for problems with large data sets. VI turns  the inference problem to an optimization problem by positing a simpler family of distributions and finding the member of the family that is closest to the true posterior distribution~\cite{wainwright2008graphical}. Hence, the inference task boils down to an optimization problem of a non-convex objective function. This allows us to bring sophisticated tools from optimization literature to tackle the performance problems. Recently, stochastic optimisation has been applied to VI in order to cope with massive data~\cite{hoffman2013stochastic}. While VI requires repeatedly iterating over the whole data set before updating the variational parameters (parameters of the variational objective), stochastic variational inference 
 (SVI) updates the parameters every time a single data example is processed. Therefore, by the end of one pass through the dataset, the parameters will have been updated multiple times. Hence, the model parameters converge faster, while using less computational resources. The idea of SVI is to move the variational parameters at each iteration  in the direction of a noisy estimate of the variational objective's natural gradient based on a couple of examples~\cite{hoffman2013stochastic}. Following these stochastic gradients with certain conditions on the (decreasing) learning rate schedule, SVI provably converges to a local optimum~\cite{robbins1951stochastic}. 
 
Although stochastic optimization improves the performance of VI, its serial employment prevents scaling up the inference and harnessing distributed resources. Since, SVI is at its core a stochastic gradient-based optimisation algorithm, horizontal parallelism is straightforward. That is, computing stochastic gradients of a batch of data samples can be done locally in parallel given that the parameters update is synchronised. Such synchronisation limits the scalability by requiring slaves to send their stochastic gradients to the master prior to each parameter update. Hence, synchronous methods suffer from the curse of the last reducer; that is, a single slow slave can dramatically slow down the whole performance. Thus, asynchronous parallel optimization is an interesting alternative provided it maintains comparable convergence rate to its synchronous counterpart. Indeed, asynchronous parallel stochastic gradient-based optimisation algorithms have recently received broad attention~\cite{recht2011hogwild,agarwal2011distributed,zhang2014asynchronous,feyzmahdavian2016asynchronous,
mania2015perturbed}. 

Authors in~\cite{agarwal2011distributed} show that for smooth stochastic convex problems the asynchronisation effects are asymptotically negligible and order-optimal convergence results can be achieved. Since, the resulting objective function of the SVI is non-convex, we are particularly interested in the asynchronous parallel stochastic gradient algorithm (ASYSG) for smooth non-convex optimization~\cite{bertsekas1989parallel}. A recent study~\cite{lian2015asynchronous} breaks the usual convexity assumption taken by~\cite{agarwal2011distributed}. Nonetheless, theoretical guarantees (convergence and speed-up) for many recent successes of ASYSG are provided. In this paper, we use the ASYSG algorithm proposed in~\cite{agarwal2011distributed} to come up with an asynchronous stochastic variational inference (ASYSVI) algorithm for a wide family of Bayesian models. We also adapt the theoretical studies of ASYSG for smooth non convex optimization from~\cite{lian2015asynchronous} to explain ASYSVIs' convergence and speed-up properties. This paper proposes a novel contribution that allows to linearly speeding up SVI by distributing its  stochastic natural gradient computations in an asynchronous way while guaranteeing an ergodic convergence rate $O(1/\sqrt{T})$ under some assumptions. We take latent Dirichlet allocation (LDA) as a case study to empirically evaluate ASYSVI.

The rest of the paper is structured as follows. We briefly review variational and stochastic variational
inference in Sec.~\ref{sec3}. We describe our asynchronous stochastic variational inference algorithm along with its convergence analysis in Sec.~\ref{sec4}. Latent Dirichlet allocation case study is developed in Sec.~\ref{sec44}. Related work is discussed in Sec.~\ref{sec5}. Empirical evaluation is presented in Sec.~\ref{sec6} and the paper concludes with a discussion in Sec.~\ref{sec7}.
\section{Background} \label{sec3}
In the following, we  derive the model family studied in this paper and review SVI. We follow the same pattern in \cite{hoffman2013stochastic}.

\textbf{Model family.} Our family of models consists of three random variables: observations $\boldsymbol x=\boldsymbol{x}_{1:n}$, local hidden variables $\boldsymbol z=\boldsymbol{z}_{1:n}$, global  hidden variables $\boldsymbol\beta$ and fixed parameters $\boldsymbol\alpha$. The model assumes that the distribution of  the $n$ pairs of $(\boldsymbol{x}_i,\boldsymbol{z}_i)$ is conditionally independent given $\boldsymbol\beta$. Further, their distribution and the prior distribution of $\boldsymbol\beta$ are in an exponential family:
\begin{equation}\label{equ1}
p(\boldsymbol\beta,\boldsymbol x,\boldsymbol z|\boldsymbol\alpha)=p(\boldsymbol\beta|\boldsymbol\alpha)\prod_{i=1}^n p(\boldsymbol{z}_i,\boldsymbol{x}_i|\boldsymbol\beta),
\end{equation}
\begin{equation}\label{equ2}
p(\boldsymbol{z}_i,\boldsymbol{x}_i|\boldsymbol\beta)=h(\boldsymbol{x}_i,\boldsymbol{z}_i)\exp\big(\boldsymbol\beta^Tt(\boldsymbol x_i,\boldsymbol z_i)-a(\boldsymbol\beta)\big),
\end{equation}
\begin{equation}\label{equ3}
p(\boldsymbol\beta|\boldsymbol\alpha)=h(\boldsymbol\beta)\exp\big(\boldsymbol\alpha^Tt(\boldsymbol\beta)-a(\boldsymbol\alpha) \big)
\end{equation}
Here, we overload the notation for the base measures $h(.)$, sufficient statistics $t(.)$ and log normalizer $a(.)$. While the soul of the proposed approach is generic, for simplicity we assume a conjugacy relationship between $(\boldsymbol x_i,\boldsymbol z_i)$ and $\boldsymbol\beta$. That is, the distribution $p(\boldsymbol\beta|\boldsymbol x,\boldsymbol z)$ is in the same family as the prior $p(\boldsymbol\beta|
\boldsymbol\alpha)$.

Note that this innocent looking family of models includes (but is not limited to) latent Dirichlet allocation \cite{blei2003latent}, Bayesian Gaussian mixture, probabilistic matrix factorization, hidden Markov models, hierarchical  linear and probit regression, and many Bayesian non-parametric models.  

\textbf{Mean-field variational inference.} Variational inference (VI) approximates intractable posterior $p(\boldsymbol\beta,\boldsymbol z|\boldsymbol x)$ by positing a family of simple distributions $q(\boldsymbol\beta,\boldsymbol z)$ and find the member of the family that is closest to the posterior (closeness is measured with KL divergence). The resulting optimization problem is equivalent maximizing the evidence lower bound (ELBO):
\begin{equation}\label{equ4}
\mathcal{L}(q)=E_q[\log p(\boldsymbol x,\boldsymbol z,\boldsymbol\beta)]-E_q[\log p(\boldsymbol z\boldsymbol\beta)]\leq \log p(\boldsymbol x)
\end{equation}
 Mean-field is the simplest family of distribution, where the distribution over the hidden variables factorizes as follows:
 \begin{equation}\label{equ5}
 q(\boldsymbol\beta,\boldsymbol z)=q(\boldsymbol\beta|\boldsymbol\lambda)\prod_{i=1}^np(\boldsymbol z_i|\boldsymbol\phi_i)
 \end{equation}
 Further, each variational distribution is assumed to come  from the same family of the true one. Mean-field variational inference optimizes the new ELBO with respect to the  local and global variational parameters $\boldsymbol\phi$ and $\boldsymbol\lambda$:
 \begin{equation}\label{equ6}
\mathcal{L}(\boldsymbol\lambda,\boldsymbol\phi)=E_q\bigg[\log\frac{p(\boldsymbol \beta)}{q(\boldsymbol \beta)}\bigg]+\sum_{i=1}^nE_q\bigg[\log\frac{p(\boldsymbol x_i,\boldsymbol z_i|\boldsymbol\beta)}{q(\boldsymbol z_i)} \bigg]
 \end{equation}
  It iteratively updates each variational parameter holding the other parameters fixed. With the assumptions taken so far, each update has  a closed form solution. The local parameters are a function of the global parameters:
 \begin{equation}\label{equ7}
 \boldsymbol\phi({\boldsymbol\lambda}_t)=\arg\max_{\boldsymbol\phi}\mathcal{L}(\boldsymbol\lambda_t,\boldsymbol\phi)
\end{equation}  
  We are interested in the global parameters which summarises the whole dataset (clusters in Bayesian Gaussian mixture, topics in LDA):
 \begin{equation}\label{equ8}
 \mathcal{L}(\boldsymbol\lambda)=\max_{\boldsymbol\phi} \mathcal{L}(\boldsymbol\lambda,\boldsymbol\phi)
\end{equation}  
To find the optimal value of $\boldsymbol\lambda$ given that $\boldsymbol\phi$ is fixed, we compute the natural gradient of $\mathcal{L}(\boldsymbol\lambda)$  and set it to zero by setting:
 \begin{equation}\label{equ9}
\boldsymbol\lambda^*=\boldsymbol\alpha +\sum_{i=1}^nE_{\boldsymbol\phi_i({\boldsymbol\lambda}_t)}[t(\boldsymbol x_i,\boldsymbol z_i)]
 \end{equation} 
Thus, the new optimal global parameters are $\boldsymbol\lambda_{t+1}=\boldsymbol\lambda^*$.  The algorithm works by iterating between computing  the optimal local parameters  given the global ones \big(Eq.~\eqref{equ7}\big) and computing the optimal global parameters given the local ones \big(Eq.~\eqref{equ9}\big).
 
\textbf{Stochastic variational inference.} Rather than analysing  all the data to compute $\boldsymbol\lambda^*$ at each iteration, stochastic optimization can be used. Assuming that the data is uniformity at random selected from the dataset, an unbiased noisy estimator of $\mathcal{L}(\boldsymbol\lambda,\boldsymbol\phi)$  can be developed based on a single data point: 
\begin{equation}\label{equ10}
\mathcal{L}_i(\boldsymbol\lambda,\boldsymbol\phi_i)=E_{q}\bigg[\log\frac{p(\boldsymbol \beta)}{q(\boldsymbol \beta)}\bigg]+nE_q\bigg[\log\frac{p(\boldsymbol x_i,\boldsymbol z_i|\boldsymbol\beta)}{q(\boldsymbol z_i)} \bigg]
\end{equation}\label{equ11}
The unbiased stochastic approximation of the ELBO as a function of $\boldsymbol\lambda$ can be written as follows:
\begin{equation}\label{equ11b}
\mathcal{L}_i(\boldsymbol\lambda)=\max_{\boldsymbol\phi_i}\mathcal{L}_i(\boldsymbol\lambda,\boldsymbol\phi_i)
\end{equation}
Following the same step in the previous section, we end up with a noisy unbiased estimate of Eq.~\eqref{equ9}: 
\begin{equation}\label{equ12}
\boldsymbol{\hat{\lambda}}=\boldsymbol\alpha +nE_{\boldsymbol\phi_i({\boldsymbol\lambda}_t)}[t(\boldsymbol x_i,\boldsymbol z_i)]
\end{equation}
At each iteration, we move the global parameters a step-size $\rho_t$ (learning rate) in the direction of the noisy natural gradient:
 \begin{equation}\label{equ13}
\boldsymbol\lambda_{t+1}=(1-\rho_t)\boldsymbol\lambda_t+\rho_t\boldsymbol{\hat{\lambda}}
\end{equation}
With certain conditions on $\rho_t$, the algorithm converges ($\sum_{t=1}^\infty\rho_t=\infty$, $\sum_{t=1}^\infty \rho_t^2<\infty$ )\cite{robbins1951stochastic}.
\section{Asynchronous Stochastic Variational Inference}\label{sec4}
In this section, we describe our proposed parallel implementation of ASYSVI on computer cluster and study its convergence and speed-up properties. The steps of the algorithm follows from the original ASYSG in~\cite{agarwal2011distributed}.

 \subsection{Algorithm Description}
 \begin{algorithm*}[!t]
   \caption{ASYSVI-Master}\label{alg1}
\begin{algorithmic}[1]
   \STATE {\bfseries Input:} number of iteration $T$ and step-size $\{\rho_t\}_{t=0,...,T-1}$
       \STATE \textbf{initialize:} $\boldsymbol\lambda^0$ randomly and $t$ to $0$
       \WHILE{($t<T$)}
        \STATE Aggregate $M$ stochastic natural gradients $\hat\nabla \mathcal{L}_1(\boldsymbol\lambda^{t-\tau_{t,1}}),...,\hat\nabla \mathcal{L}_M(\boldsymbol\lambda^{t-\tau_{t,M}})$ $\boldsymbol$ from the slaves\STATE Average the $M$ stochastic natural gradients. $G^t_M=\sum_m \hat\nabla \mathcal{L}_m(\boldsymbol\lambda^{t-\tau_{t,m}})$ 
      \STATE Update the current estimate of the global variational parameter. $\boldsymbol \lambda^{t+1}=\boldsymbol\lambda^t+\rho_t G^t_M$
      \STATE $t=t+1$
     \ENDWHILE
\end{algorithmic}
\end{algorithm*}
\begin{algorithm*}[!t]
   \caption{ASYSVI-Slave}\label{alg2}
\begin{algorithmic}[1]
 \STATE {\bfseries Input:} data size $D$  
       \WHILE{(True)}
        \STATE Sample a data point $\boldsymbol x_i$ uniformly from the data set 
        \STATE Pull a global variational parameter $\boldsymbol \lambda^*$ from the master
         \STATE Compute the local variational parameters $\boldsymbol\phi_i^*({\boldsymbol\lambda^*})$ corresponding to the  data point $\boldsymbol x_i$ and the global variational parameter $\boldsymbol \lambda^*$, $\boldsymbol\phi_i^*({\boldsymbol\lambda^*})=\arg\max_{\boldsymbol\phi_i}\mathcal{L}_i(\boldsymbol\lambda^*,\boldsymbol\phi_i)$\STATE Compute the stochastic natural gradient with respect to the global parameter $\boldsymbol\lambda$, $\boldsymbol g_i(\boldsymbol\lambda)=\boldsymbol\alpha +DE_{\boldsymbol\phi_i({\boldsymbol\lambda})}[t(\boldsymbol x_i,\boldsymbol z_i)]-\boldsymbol\lambda$  \STATE Push $\boldsymbol g_i(\boldsymbol\lambda^*)$ to the master
     \ENDWHILE
\end{algorithmic}
\end{algorithm*}
 ASYSVI is presented analogously to ASYSG in \cite{lian2015asynchronous} but in the context of VI. The architecture of the computer network on which ASYSVI is run is known as the \textit{star-shaped} network. In this network, a master machine maintains the global variational parameter $\boldsymbol \lambda$, whereas other machines serve as slaves which independently and simultaneously compute the local variational parameters $\boldsymbol\phi$ and stochastic gradients of ELBO $\mathcal{L}(\boldsymbol\lambda)$. The slaves only communicate with the master to exchange information in which they access the state of the global variational parameter and provide the master with the stochastic gradients. These gradients are computed with respect to $\boldsymbol \lambda$ based on few (mini-batched or single) data points acquired from distributed sources. The master aggregates predefined amounts of stochastic gradients from slaves nonchalantly about the sources of the collected stochastic gradients. Then, it updates its current global variational parameter. The update step is performed as an atomic operation where slaves cannot read the value of the global variational parameter during this step. However, vertical parallelism can be achieved by adopting the ASYSG algorithm proposed in~\cite{recht2011hogwild}. Furthermore, a hybrid horizontal-vertical parallelism could be achieved by combining the mechanism used in~\cite{raman2016extreme} with ASYSVI (more details in Sec.~\ref{sec5} and Sec.~\ref{sec7}) 

The key difference between ASYSVI and the  synchronous parallel SVI is that ASYSVI does not lock the slaves until the master's update step is done. That is, the slaves might compute some stochastic gradients based on early value of the global variational parameter. By allowing delayed and asynchronous updates, one might expect slower convergence if any. In the next section, we apply the study of~\cite{lian2015asynchronous} on SVI to show that the effect of stochastic gradients delay will vanish asymptotically. The algorithms of ASYSVI-mater and ASYSVI-salve are shown in Alg.~\ref{alg1} and Alg.~\ref{alg2}. We denote by $\tau_{t,m}$ the delays between the current iteration $t$ and the one when the slave pulled the global variational parameter at which it computed the stochastic gradient. \vspace{-0.2cm}
\subsection{Convergence Analysis}
Following~\cite{lian2015asynchronous,agarwal2011distributed}, we take the same assumptions, but replace the gradient with the natural gradient:
\begin{itemize}
\item Unbiased gradient: the expectation of the stochastic natural gradient of Eq.~\eqref{equ11b} is equivalent to the natural gradient of Eq.~\eqref{equ8}: 
\begin{equation}
\hat\nabla \mathcal{L}(\boldsymbol\lambda)=E[\hat\nabla\mathcal{L}_i(\boldsymbol\lambda)]
\end{equation} 
where $\hat\nabla$ denotes natural gradient. This assumption already holds in SVI problems for the family of models shown in Sec.~\ref{sec3}. 
\item Bounded variance: the variance of the stochastic natural gradient is bounded for all $\lambda \in\mathcal{G}$, $E[||\hat\nabla\mathcal{L}_i(\boldsymbol\lambda)-\hat\nabla \mathcal{L}(\boldsymbol\lambda)||^2]\leq \sigma^2$. By applying SVI natural gradient, we end up with the following formulation:
\begin{equation}
E[||nE_{\boldsymbol\phi_i({\boldsymbol\lambda})}[t(\boldsymbol x_i,\boldsymbol z_i)]-\sum_{i=1}^nE_{\boldsymbol\phi_i({\boldsymbol\lambda})}[t(\boldsymbol x_i,\boldsymbol z_i)]||^2]\leq \sigma^2
\end{equation}
\item Lipschitz-continuous gradient: the natural gradient is L-Lipschitz-continuous for all $\lambda\in\mathcal{G}$ an $\lambda'\in \mathcal{G}$, $||\hat\nabla\mathcal{L}(\boldsymbol\lambda)-\hat\nabla \mathcal{L}(\boldsymbol\lambda')||\leq L||\lambda-\lambda'||$. By applying SVI natural gradient, we end up with the following formulation: 
\begin{equation}||\sum_{i=1}^nE_{\boldsymbol\phi_i({\boldsymbol\lambda})}[t(\boldsymbol x_i,\boldsymbol z_i)]-\lambda-\sum_{i=1}^nE_{\boldsymbol\phi_i({\boldsymbol\lambda'})}[t(\boldsymbol x_i,\boldsymbol z_i)]+\lambda'||\leq L||\lambda-\lambda'||
\end{equation}
\item Bounded delay: All delay variables $\tau_{t,m}$ are bounded: 
\begin{equation}
\max_{t,m}\tau_{t,m}\leq B
\end{equation}
\end{itemize} 
In addition to these assumptions, authors~\cite{lian2015asynchronous,agarwal2011distributed} assume that each slave receives a stream of independent data points. Although this assumption might not be satisfied strictly in practice, we follow the same assumption for analysis purpose. Thus, the same theoretical results obtained by~\cite{lian2015asynchronous} can be applied for ASYSVI, namely, an ergodic convergence rate $O(1/\sqrt{MT})$ provided that $T$ is greater than $O(B^2)$. The results also show that, since the number of slaves is proportional to $B$, the ergodic convergence rate is achieved as long as the number of salves is bounded by $O(\sqrt{T/M})$. Note that $O(1/\sqrt{MT})$ is consistent with the serial stochastic gradient (SG) and the stochastic variational inference (SVI). Thus, ASYSG and ASYSVI allow for a linear speed-up if $B\leq O(\sqrt{T/M})$.
\section{Case Study: Latent Dirichlet Allocation}\label{sec44}
 Latent Dirichlet allocation (LDA) is an instance of the family of models described in Sec~\ref{sec3} where the global, local, observed variables and their distributions are set as follows:
\begin{itemize}
\item the global variables $\{\boldsymbol\beta\}_{k=1}^K$ are the topics in LDA. A topic is a distribution over the vocabulary, where the probability of a word $w$ in topic $k$ is denoted by $\beta_{k,w}$. Hence, the prior distribution of $\boldsymbol\beta$ is a Dirichlet distribution $p(\boldsymbol\beta)=\prod_kDir(\boldsymbol\beta_k;\boldsymbol\eta)$
\item the local variables are the topic proportions $\{\boldsymbol\theta_d\}_{d=1}^D$ and the topic assignments $\{\{z_{d,w}\}_{d=1}^D\}_{w=1}^{W}$ which index the topic that generates the observations. Each document is associated with a topic proportion which is a distribution over topics, $p(\boldsymbol\theta)=\prod_dDir(\boldsymbol\theta_d;\boldsymbol\alpha)$. The assignments  $\{\{z_{d,w}\}_{d=1}^D\}_{w=1}^{W}$  are indices, generated by $\boldsymbol\theta_d$, that couple topics with words, $p(\boldsymbol z_d|\boldsymbol\theta)=\prod_w\theta_{d,z_{d,w}}$
\item the observations $\boldsymbol x_{d}$ are the words of the documents which are assumed to be drawn from topics $\boldsymbol\beta$ selected by indices $\boldsymbol z_{d}$ , $p(\boldsymbol x_d|\boldsymbol z_d,\boldsymbol\beta)= \prod_{w}\beta_{z_{d,w},x_{d,w}}$
\end{itemize}
 The basic idea of LDA is that documents are represented as random mixtures over latent topics, where each topic is characterized by a distribution over words~\cite{blei2003latent}. LDA assumes the following generative process:
\begin{itemize}
\item[1]Draw topics $\boldsymbol\beta_k\sim Dir(\eta,...,\eta)$ for $k\in\{1,...,K\}$
\item[2]Draw topic proportions $\boldsymbol\theta_d\sim Dir(\alpha,...,\alpha)$ for $d\in\{1,...,D\}$
\begin{itemize}
\item[2.1]Draw topic assignments $z_{d,w}\sim Mult(\boldsymbol\theta_d)$ for $w\in\{1,...,W\}$
\begin{itemize}
\item[2.1.1] Draw word $x_{d,w}\sim Mult(\boldsymbol\beta_{z_{d,w}})$
\end{itemize}
\end{itemize}
\end{itemize}
According to Sec.~\ref{sec3}, each variational distribution is assumed to come  from the same family of the true one. Hence, $q(\boldsymbol\beta_k|\boldsymbol\lambda_k)=Dir(\boldsymbol\lambda_k)$, $q(\boldsymbol\theta_d|\boldsymbol\gamma_d)=Dir(\boldsymbol\gamma_d)$ and $q(z_{d,w}|\boldsymbol\phi_{d,w})=Mult(\boldsymbol\phi_{d,w})$. To compute the  stochastic natural gradient $g_i$ in Alg.~\ref{alg2} for LDA, we need to find the sufficient statistic $t(.)$ presented in Eq.~\eqref{equ2}. By writing the likelihood of LDA in the form of Eq.~\eqref{equ2}, we obtain $t(\boldsymbol x_d, z_d)=\sum_{w=1}^W\mathbf{I}_{z_{d,w},x_{d,w}}$, where $\mathbf{I}_{i,j}$ is equal to $1$ for entry $(i, j)$ and $0$ for all the rest. Hence, the stochastic natural gradient $g_i(\boldsymbol\lambda_k)$ can be written as follows:
\begin{equation}\label{equ14}
g_i(\boldsymbol\lambda_k)=\eta+D\sum_{w=1}^W\phi_{i,w}^k\mathbf{I}_{k,x_{i,w}}-\boldsymbol\lambda_k
\end{equation}
Details on how to compute the local variational parameters $\boldsymbol\phi_i^*({\boldsymbol\lambda^*})$ in Alg.~\ref{alg2} can be found in~\cite{hoffman2013stochastic}. 

Having computed the elements needed to run ASYSVI's algorithms~\ref{alg1} and~\ref{alg2}, we move to the convergence analysis. Since the data is assumed to be subsampled uniformly, the gradient unbiased assumption holds for LDA. We can always find a constant variable such that the bounded variance is satisfied. At the worst case, the variance of the stochastic natural gradient of LDA can be bounded by $DW\big( max_{i,w}(\phi^k_{i,w})^2-min_{i',w'}(\phi^k_{i',w'})^2\big)$, $\forall k$. Therefore, it can be bounded by $O((DW)^2)$. It is clear that the Lipschitz-continuous gradient can be satisfied for any class of the family models proposed in Sec.~\ref{sec3} and hence, for LDA. Finally, the bounded delay can be guaranteed through the implementation. Therefore, ASYSVI of LDA can converge since the aforementioned assumptions can be satisfied. 
\section{Related Work}\label{sec5}
Few work has been proposed to scale variational inference to large datasets. We can distinguish two major classes. The first class is based on the Bayesian filtering approach~\cite{honkela2003line,broderick2013streaming}. That is, the sequential nature of Bayes theorem is exploited to recursively update an approximation of the posterior. Particularly, variational inference  is used between the updates to approximate the posterior which becomes the prior of the next step. Author in~\cite{honkela2003line} employs forgetting factors to decay the contributions from old data points in favour of a new better one. The algorithm proposed in~\cite{broderick2013streaming} considers a sequence of data batches. It iterates over the data points in the batch until convergence. The computation of the batches posterior is done in a distributed and asynchronous manner. That is, the algorithm applies VI by performing asynchronous Bayesian updates to the posterior as batches of data arrive continuously. Similar to our approach, master-slave architecture is used.

The second class of work is based on optimization~\cite{hoffman2013stochastic,neiswanger2015embarrassingly,raman2016extreme}. As we already discussed, SVI proposed by~\cite{hoffman2013stochastic} employs stochastic optimization to scale up Bayesian computation to massive data. SVI is inherently serial and requires the model parameters to fit in the memory of a single processor. Authors in~\cite{neiswanger2015embarrassingly} presents an inference algorithm, where the data is divided across several slaves and each of them perform VI updates in parallel. However at each iteration, the slaves are synchronized to combine their obtained parameters. Such synchronisation limits the scalability and decreases the speed of the update to that of the slowest slave. To avoid  bulk synchronization, authors in~\cite{raman2016extreme} propose an asynchronous and lock-free update. In this update, vertical parallelism is adopted, where each processor asynchronously updates a subset of the parameters based on a subset of the data attributes. In contrast, we adopt horizontal parallelism update based on few (mini-batched or single) data points acquired from distributed sources. The update steps are, then, aggregated to form the global update. Note that the proposed approach can make use of the mechanism proposed by~\cite{raman2016extreme} to achieve  a hybrid horizontal-vertical parallelism. On the contrary to~\cite{raman2016extreme}, our approach is not customised for LDA and can be simply applied to any model from the family of models presented in Sec.~\ref{sec3}
\section{Experimental Results}\label{sec6}
\begin{table*}[t]
\centering
\caption{Parameters settings}
\label{my-label}
\begin{adjustbox}{width=1\textwidth}
\begin{tabular}{|l|l|l|l|l|l|l|l|l|l|l|l|l|l|l|l|l|l|}
\hline
Data sets & \multicolumn{4}{l|}{Enron emails}& \multicolumn{4}{l|}{NYTimes news articles}&   \multicolumn{4}{l|}{Wikipedia articles}  \\\hline
$batch$          & 16 & 64 & 256 &  1024& 16 & 64 & 256 &  1024 & 16 & 64 & 256 &  1024   \\ \hline
$\kappa$    & 0.7 & 0.7 & 0.5 & 0.5& 0.7 & 0.7 & 0.5 &  0.5 &0.7&0.7&0.5&0.5  \\\hline
$\tau_0$    & 1024 & 24 & 24 &  1& 1024 & 24 & 24 & 1 &1024&1024&1024&1024 \\ \hline
perplexity  &5919 & 5348 & 5264 & 4771 &  11989 & 10156 & 9015 &5501 & 1446& 1390&1355&  1332 \\ \hline
\end{tabular}
\end{adjustbox}
\end{table*}    
\begin{figure*}[t]
\subfigure[TSP]{\includegraphics[width = 0.498\textwidth, height=6cm]{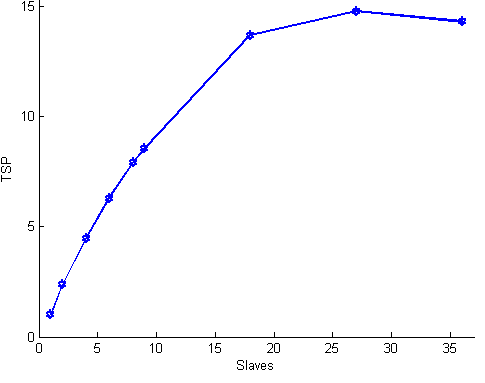}}
\subfigure[RSP]{\includegraphics[width = 0.498\textwidth, height=6cm]{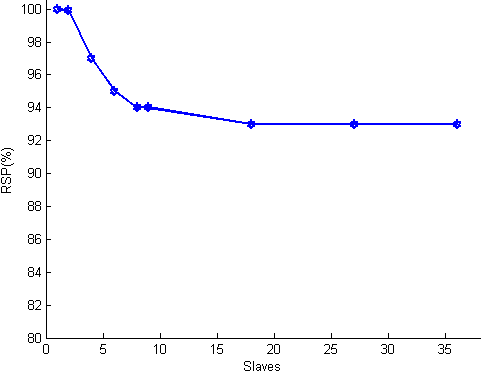}}
\caption{Comparing ASYSVI LDA to online LDA on \textit{Enron} dataset}\label{fige4}
\end{figure*} 
 \begin{figure*}[t]
 \subfigure[TSP on \textit{Enron} dataset]{\includegraphics[width = 0.498\textwidth, height=6cm]{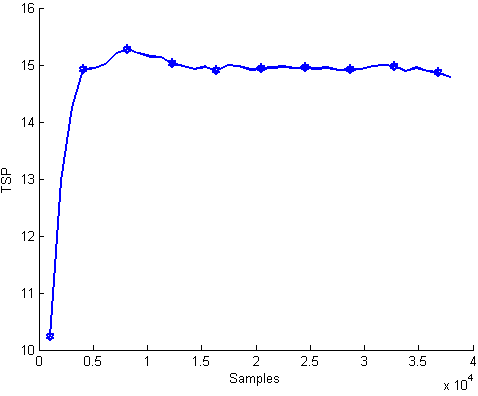}}
\subfigure[RSP on \textit{Enron} dataset]{\includegraphics[width = 0.498\textwidth, height=6cm]{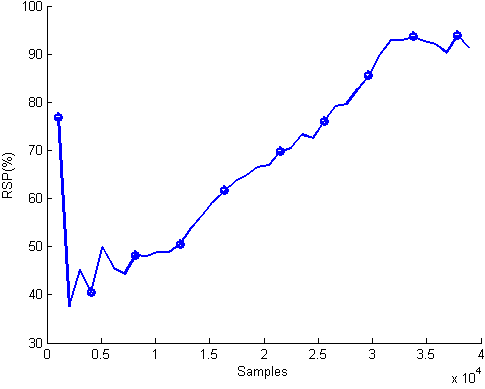}}
\caption{TSP and RSP with respect to streamming samples on \textit{Enron} dataset}\label{fige5}
\vspace{-0.2cm}\end{figure*}
 \begin{figure*}[t]
\subfigure[TSP on \textit{NYTimes} dataset]{\includegraphics[width = 0.498\textwidth, height=6cm]{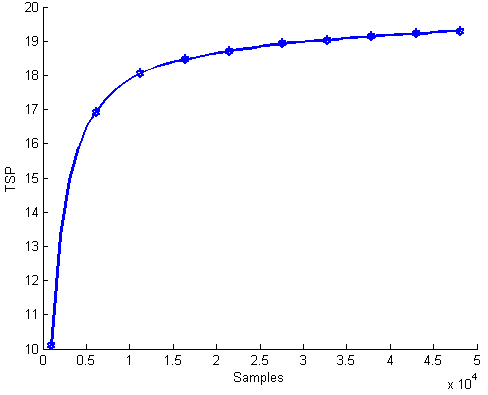}}
\subfigure[RSP on \textit{NYTimes} dataset]{\includegraphics[width = 0.498\textwidth, height=6cm]{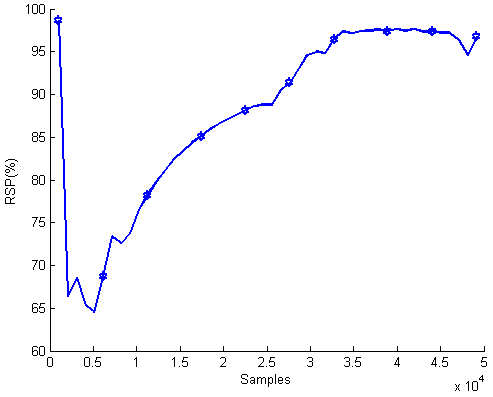}}
\caption{TSP and RSP with respect to streamming samples on \textit{NYTimes} dataset}\label{fige6}
\end{figure*}
 \begin{figure*}[t]
\subfigure[TSP on \textit{Wikipedia} dataset]{\includegraphics[width = 0.498\textwidth, height=6cm]{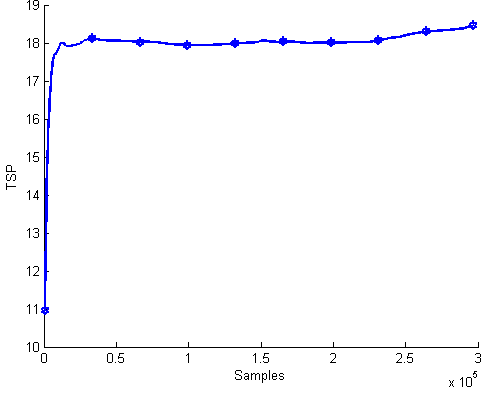}}
\subfigure[RSP on \textit{Wikipedia} dataset]{\includegraphics[width = 0.498\textwidth, height=6cm]{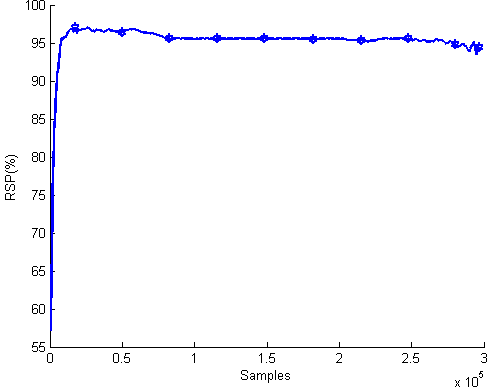}}
\caption{TSP and RSP with respect to streamming samples on \textit{Wikipedia} dataset}\vspace{-0.2cm}\label{fige7}
\end{figure*}
In the following, we demonstrate the usefulness of distributing the computation of SVI, mainly the speed-up advantages of ASYSVI. For this purpose, we compare the speed-up of ASYSVI LDA against serial SVI LDA (online LDA~\cite{hoffman2010online}). The two versions of LDA are evaluated on three datasets consisting of very large collections of documents. We also evaluate ASYSVI LDA in the streaming setting where new documents arrive in the form of stream. The algorithm implementation is available in Python~\footnote{Code will be uploaded later}. 

The performance evaluation is done using held-out perplexity as a measure of model fit. Perplexity is defined as the geometric mean of the inverse marginal probability of each word in the held-out set of documents~\cite{blei2003latent}. To validate the speed-up properties, following~\cite{lian2015asynchronous}, we compute the running time speed-up (TSP): 
\begin{align}
TSP&=\frac{running\quad time\quad  of\quad online\quad LDA}{running\quad time\quad  of\quad asynchronous\quad LDA}
\end{align}
\textbf{Datasets:} we perform all comparisons and evaluations on three corpora of documents. The first two corpora are available on~\cite{Lichman:2013}. The third corpus was used in~\cite{hoffman2010online}.
\begin{itemize}
\item \textit{Enron} emails: The corpus contains $39,861$ email messages from about $150$ users, mostly senior management of Enron. Data is proceeded before usage by removing  all words not in a vocabulary dictionary of $28,102$ words.
\item \textit{NYTimes} news articles: The corpus contains $300,000$ news articles from the \textit{New York Times}. Data is proceeded before usage by removing  all words not in a vocabulary dictionary of $102,660$ words.
\item \textit{Wikipedia} articles: this corpus contains $1M$ documents downloaded from \textit{Wikipedia}. Data is proceeded before usage by removing  all words not in a vocabulary dictionary of $7,700$ words.
\end{itemize}

\textbf{Settings the parameters:} In all experiments, $\alpha$ and $\eta$ are fixed at $0.01$ and the number of topics $K = 50$. We evaluated a range of settings of the learning parameters, $\kappa$, $\tau_0$, and $batch$ on all four corpora. The parameters $\kappa$ and $\tau_0$, defined in~\cite{hoffman2010online}, control the learning steps-size $\rho_t$. For each corpora, we use $29,861$ emails from \textit{Enron} dataset, $50,000$ news articles from \textit{NYTimes} dataset and $300,000$ documents from \textit{Wikipedia} dataset as training sets. We also reserve $5,000$ documents as a validation set and another $5,000$ documents as a testing set. The online LDA  is run (one time per corpus) on the training sets of each corpus for $\kappa\in \{0.5, 0.7, 0.9\}$, $\tau_0\in\{1, 24, 256, 1024\}$, and $batch\in \{16, 64, 256, 1024\}$. Table~\ref{my-label} summarises the best settings of each $batch$ along with the resulting perplexity on the test set for each corpus.
    
\textbf{Comparing Serial online LDA and asynchronous LDA:} for each dataset, we set the parameters setting that give the best performance (least perplexity). ASYSVI LDA is then compared against serial SVI LDA using the same parameters setting. 

The code is implemented on high-performance computing (HPC) environment using  message passing interface (MPI) for python (MPI4py). The cluster consists of 10 nodes, excluding the head node, with each node is a four-core processor. We run AYSVI LDA on \textit{Enron} dataset for number of workers $nW\in \{2, 4, 6, 8 ,9, 18, 27, 36\}$, $B$ is set to $5$. The number of employed nodes is equal to $nW$ as long as $nW$ is less than $9$. As $nW$ becomes higher than the available nodes, the processors' cores of nodes are employed as slaves until all the cores of all the nodes are used i.e., $9\times 4=36$. Since the batch size is fixed to $1024$, each slave processes a batch of data of size $S=1024/M$ per iteration, where $M$ is fixed to $36$. Thus, the gradient computed by each slave will be multiplied by $D/S$. Hence, line number $6$ of Alg.~\ref{alg2} becomes, $\boldsymbol g_i(\boldsymbol\lambda)=\boldsymbol\alpha +(D/S)E_{\boldsymbol\phi_i({\boldsymbol\lambda})}[t(\boldsymbol x_i,\boldsymbol z_i)]-\boldsymbol\lambda$. 

 Figure~\ref{fige4} summarises the total speed up (i.e., TSP measured at the end  of the algorithm)  as well as the ratio of serial LDA perplexity to parallel LDA (RSP) on the test set for \textit{Enron} dataset. It shows TSP and RSP results with respect to the number of slaves. It is clear that as long as each node is assigned one slave, the speed-up is linear which demonstrates the convergence analysis done in Sec.~\ref{sec4}. Linear speed-up slowly converts to sub-linear as solo machines host more than one slave. The main reason of such behaviour is the communication delay caused by the increase of the network traffic. Hence, TSP is affected by the hardware. The communication cost starts affecting the algorithm speed-up when it becomes comparable to the local computation. Hence, increasing the local computation by increasing the batch size can be adopted to soften the communication effect. However, this decreases the convergence rate and increase local memory load. Hence, a balanced trade-off should be considered. RSP in Figure~\ref{fige4} shows that although the speed of online LDA has been increased up to 15 times, performance is not seriously affected. We also evaluate TSP and RSP on \textit{NYTimes} and \textit{Wikipedia} for $nW=27$. The processing speed of Online LDA on \textit{NYTimes} has been increased $19.29$ times, $TSP=19.29$, with slight loss of performance, $RSP=0.97$. For \textit{Wikipedia}, $TSP=18.58$ and $RSP=0.94$. 

Figures~\ref{fige5}, \ref{fige6} and \ref{fige7} present TSP and  RSP with respect to streaming samples from \textit{Enron}, \textit{NYTimes} and \textit{Wikipedia} datasets. These figures show the performance of ASYSVI in a true online setting where the algorithm continually collects samples from the hard driver for the case of \textit{Enron} and \textit{NYTimes} or by downloading online in the case of \textit{Wikipedia}. The perplexity is obtained online on the coming batches before being used to update the model parameters. The plots in the figures are slightly softened using a low-pass filter in order to make them easy to read. These plots show that the speed-up becomes invariant as more samples are processed. The poor speed-up in the beginning is normally caused by initialization and loading process. It can be noticed that the performance of ASYSVI LDA suffers at the beginning then it becomes comparable to online LDA after certain number of iterations. This behaviour can be explained by the convergence condition shown in Sec.~\ref{sec4} ($T$ is greater than $O(B^2)$). Thus, as the number of iterations increases, the convergence of  ASYSVI LDA is guaranteed and its performance becomes comparable to that of online LDA. Hence, RSP approaches $1$. 
\vspace{-0.2cm}
\section{Conclusion and Discussion}\label{sec7}

We have introduced ASYSVI, an asynchronous parallel implementations for SVI on computer cluster. ASYSVI leads to linear speed-up, while guaranteeing an asymptotic convergence rate given some assumptions involving the number of the slaves and iterations. Empirical results  using latent Dirichlet allocation topic model as a case study have demonstrated the advantages of ASYSVI over SVI, particularly with respect to the key issue of speeding-up the computation while maintaining comparable performance to SVI.

In future work, vertical parallelism can be adopted along with the proposed horizontal one leading to a hybrid horizontal-vertical parallelism. In such case, multi-core processors will be used for the vertical parallelism, while horizontal parallelism is achieved on a multi-node machine. Another avenue of interest is to derive an algorithm for streaming, distributed, asynchronous inference where the number of instances is not known. Moreover, it is interesting to apply ASYSVI on very large scale problems and particularly on other models of the family discussed in Sec.~\ref{sec3} and studying the effect of the statistical properties of those models. \vspace{-0.2cm}
\section*{Acknowledgment}
S. Mohamad and A. Bouchachia were supported by the European Commission under the Horizon 2020 Grant 687691 related to the project: PROTEUS: Scalable Online
Machine Learning for Predictive Analytics and Real-Time Interactive Visualization.\vspace{-0.25cm}
\bibliographystyle{ACM-Reference-Format}
\bibliography{sample-bibliography}

\end{document}